\documentclass{article}
\usepackage{spconf,amsmath,graphicx }
\title{UNSUPERVISED DISCOVERY OF LINGUISTIC STRUCTURE INCLUDING TWO-LEVEL ACOUSTIC PATTERNS USING THREE CASCADED STAGES OF ITERATIVE OPTIMIZATION}
\name{Cheng-Tao Chung$^{\#1}$, Chun-an Chan$^{*2}$, and Lin-shan Lee$^{\#*3}$}
\address{
Graduate Institute of Electrical Engineering, National Taiwan University$^{\#}$\\
Graduate Institute of Communication Engineering, National Taiwan University$^{*}$\\
\small
\texttt{r01921031@ntu.edu.tw$^{1}$, chunanchan@gmail.com$^{2}$, lslee@gate.sinica.edu.tw$^{3}$}
\normalsize
}

\begin{document}
\ninept
%\fontsize{10}{30}\selectfont
\maketitle
\begin{abstract}
Techniques for unsupervised discovery of acoustic patterns are getting increasingly attractive, because huge quantities of speech data are becoming available but manual annotations remain hard to acquire. In this paper, we propose an approach for unsupervised discovery of linguistic structure for the target spoken language given raw speech data. This linguistic structure includes two-level (subword-like and word-like) acoustic patterns, the lexicon of word-like patterns in terms of subword-like patterns and the N-gram language model based on word-like patterns. All patterns, models, and parameters can be automatically learned from the unlabelled speech corpus. This is achieved by an initialization step followed by three cascaded stages for acoustic, linguistic, and lexical iterative optimization. The lexicon of word-like patterns defines allowed consecutive sequence of HMMs for subword-like patterns. In each iteration, model training and decoding produces updated labels from which the lexicon and HMMs can be further updated. In this way, model parameters and decoded labels are respectively optimized in each iteration, and the knowledge about the linguistic structure is learned gradually layer after layer. The proposed approach was tested in preliminary experiments on a corpus of Mandarin broadcast news, including a task of spoken term detection with performance compared to a parallel test using models trained in a supervised way. Results show that the proposed system not only yields reasonable performance on its own, but is also complimentary to existing large vocabulary ASR systems.
\end{abstract}
\section{Introduction}
\label{sec:intro}
Supervised training of HMMs for automatic speech recognition relies on not only collecting huge quantities of acoustic data, but also obtaining the corresponding precise labels. Such supervised training method yields adequate performance in most circumstances but with high cost, and in many situations such annotated data sets are simply not available. This is why substantial effort \cite{jansen1}-\cite{chan3} has been made for unsupervised discovery of acoustic patterns from huge quantities of acoustic data which may be easily obtained nowadays, without manual labels and corresponding knowledge. Most of such effort discovered only one level of phoneme like acoustic patterns. However, it is well known that speech signals have multi-level structure including at least phoneme and words, and such structure are very helpful in analysing or decoding speech\cite{filler11}.

In this paper we propose an approach for unsupervised discovery of structured two-level acoustic patterns including subword-like patterns and word-like patterns (concatenation of several subword-like patterns). Not only the HMMs for these patterns, the number of the subword-like patterns and the lexicon size of word-like patterns can be automatically learned from data, but more knowledge about the language such as the N-gram language model and the word-like pattern lexicon, jointly referred to as the linguistic structure in this paper, can all be obtained directly from the acoustic signals of a corpus. This is achieved by integrating a dynamic lexicon into the process of the conventional supervised HMM-training, and performing three stages of iterative optimization between the labels and the models, such that the models, parameters, and the linguistic structure can then collect knowledge from the corpus layer after layer iteratively and adjust themselves accordingly. In this way, we are able to develop semantic building blocks of the target spoken language represented by the corpus with word-like patterns and acoustic building blocks of the target spoken language with subword-like patterns.

\section{Proposed approach: cascaded three stages of iterative optimization}
\label{sec:pagestyle}
The goal is to find the parameter set $\theta=\{\theta^a,\theta^x,\theta^l\}$ for the linguistic structure and the word-like pattern label $W$ given the observed acoustic feature vector sequences $\bar{O}$ for the corpus considered. The parameter set $\theta$ includes three parts: 
$\theta^a$ for acoustic HMMs of subword-like patterns, $\theta^x$ for lexicon of word-like patterns in terms of subword-like pattern sequences, and $\theta^l$ for N-gram word-like pattern language model. This is achieved by first finding an initial label $W_0$ for the observation $\bar{O}$ as in (\ref{eq:1}). In each iteration $i$, we train the parameters $\theta_i$ with the label $W_{i-1}$ obtained in the previous iteration as in (\ref{eq:2}) and decode the label $W_{i}$ with the obtained parameters $\theta_i$ as in (\ref{eq:3}).

\begin{eqnarray}
W_0 &=& \mbox{initialization}(\bar{O}),                                   \label{eq:1} \\ 
\theta_{i} &=& \arg \max_{\substack{\theta}} P(\bar{O}|\theta,W_{i-1}),   \label{eq:2} \\
W_i &=& \arg \max_{\substack{W}} P(\bar{O}|\theta_i ,W).                  \label{eq:3}
\end{eqnarray}

\begin{figure}[tbh]
  \centering
    \includegraphics[width=0.45\textwidth]{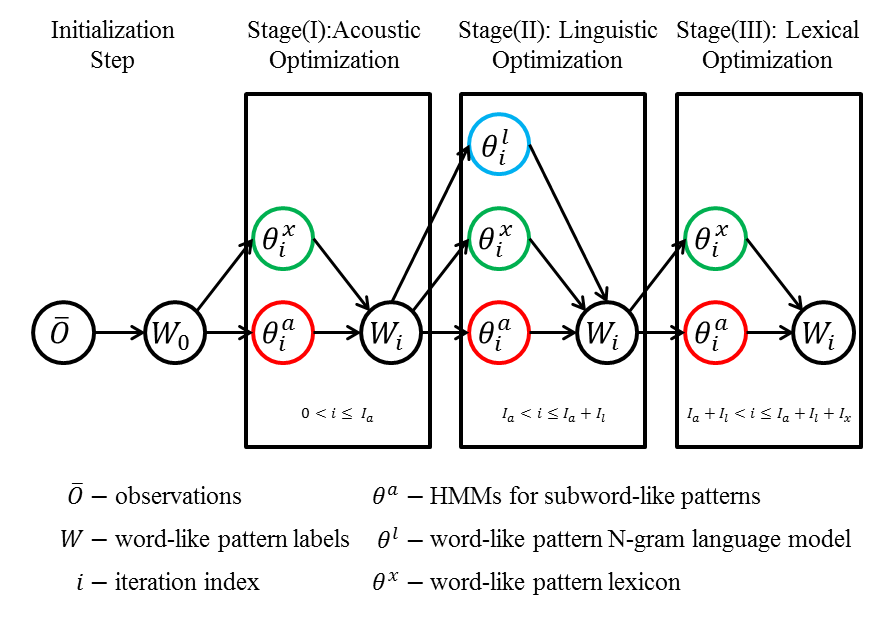}
     \caption{
	Simplified digram for the proposed initialization step followed by three stages of iterative optimization.
     The four phase iterative optimization procedure. Some dependency links have been omitted.}
     \label{fig:FP}
\end{figure}

\begin{figure}[h]
\centering
\includegraphics[width=0.40\textwidth]{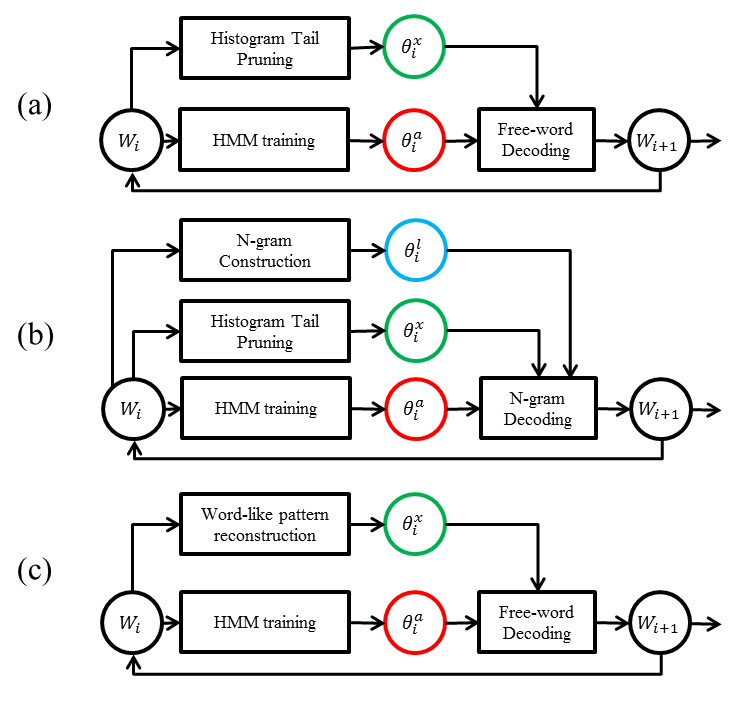}
\caption{Detailed diagrams for the three stages of (a)acoustic (b)linguistic and (c)lexical optimization.}
\label{fig:Pac}
\end{figure}

The iterations above are organized as an initialization step followed by three cascaded stages (I)(II)(III) respectively for acoustic, linguistic and lexical optimization as shown in Fig. \ref{fig:FP}. 
In Fig. \ref{fig:FP}, the number of iterations for each stage are $I_a$, $I_l$ and $I_x$ respectively. When the difference between $W_{i-1}$,  $W_{i}$ becomes insignificant, the process then advances to the next stage. The parameters $\theta^a_i$ are generated by EM training as in (\ref{eq:2}), while the other parameters $\theta^l_i$ or $\theta^x_i$ are generated directly from the labels $W_{i-1}$ obtained in the previous iteration.
However, not all of $\theta^a$, $\theta^x$, and $\theta^l$ are used in each stage. The detailed updating procedure is depicted in Fig. \ref{fig:Pac} and will be explained shortly.

The basic idea behind the procedure in Fig. \ref{fig:FP} is to gradually construct and update the parameters layer after layer. This prevents the parameters from being caught in local optimal situations which often happen when too many parameters are optimized at once.
First, the HMM parameters for the subword-like patterns are trained alone in stage (I), because these HMMs are the primary building blocks of the whole linguistic structure and reliable estimate for their parameters is the key. With reliable enough HMMs for subword-like patterns, we then in stage (II) use N-gram parameters for word-like patterns to better decode those word-like patterns frequently appearing together while continuously updating the HMM parameters. Finally in the stage (III), we break the word-like patterns into subword-like patterns and reconstruct better word-like patterns. The number of word-like patterns in the lexicon may shrink in the iterations of the first two stages because some less frequent patterns can be absorbed by other patterns, but this number can be changed significantly in the third stage. The time alignment for the subword-like patterns are updated in all iterations when the the labels $W_i$ are decoded.


\subsection{Initialization Step}
\begin{figure}[tbh]
  \centering
    \includegraphics[width=0.16\textwidth]{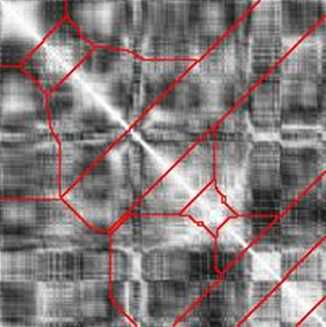}
     \caption{
An example dotplot and its watershed transform.}
     \label{fig:WS}
\end{figure}

%The initial labels do not have to be very precise, because they can be updated by many following iterations. 
Here we initialize the labels in a top-down fashion by first breaking each utterance into word-like segments based on the discontinuities in a parameter evaluated from energy and MFCC features. For each word-like segment, we further divide it into subword-like segments in the following way. We perform a watershed transform on the filtered self-similarity dotplot \cite{jansen2} for acoustic features of each hypothesized word-like segment. Watershed transformation is able to capture the number of objects and their borders in a gray scale image \cite{watershed}. So, the intersections of the diagonal entries of the dot-plot with the watershed transform object borders are taken as the boundaries between subword-like segments. An example dotplot and its watershed transform including the hypothesized subword-like segment boundaries is shown in Fig. \ref{fig:WS}. 

We then extract an average representative feature vector for every hypothesized subword-like segment, and perform global k-means clustering on these representative vectors obtained from the whole corpus. The number of clusters (the initial number of subword-like patterns) is determined by the ratio of the within-cluster total scattering to the between-cluster total scattering. A subword-like pattern ID is then assigned to each cluster. A distinct sequence of consecutive subword-like patterns for word-like segments then defines a word-like pattern, and the total number of distinct word-like patterns in the corpus is the initial vocabulary size of the lexicon. The corpus is thus represented by its initial labels $W_0$. 

\subsection{Stage(I):Acoustic Optimization}

% in terms of definition for word-like units.
The process in stage(I) is shown in Fig. \ref{fig:Pac}(a). In each iteration, the acoustic model set $\theta^a_i$ are the HMMs trained from the corpus based on $W_i$ with the ML criterion. The lexicon $\theta^x_i$ is derived by collecting all word-like patterns appearing in $W_i$ with counts exceeding a threshold. Free word decoding is then performed on the whole corpus $\bar{O}$ based on $\theta^a_i$ and $\theta^x_i$, producing an updated label $W_{i+1}$. When $W_i$ is updated to $W_{i+1}$, not only the HMM parameters of $\theta^a_i$ and HMM segmentation boundaries are updated, but the vocabulary size of $\theta^x_i$ may shrink when the counts of some word-like patterns become small enough.

\subsection{Stage(II):Linguistic Optimization}
This stage is shown in Fig. \ref{fig:Pac}(b), which is very similar to the previous stage. The only difference is an N-gram language model $\theta^l_i$ for the word-like patterns is estimated from the label $W_i$ and is used in decoding to produce the updated labels $W_{i+1}$. The N-grams help produce better labels $W_{i+1}$ especially for word-like patterns appearing together frequently. 

\subsection{Stage(III):Lexical Optimization}
We reconstruct new word-like patterns in this step as in Fig. \ref{fig:Pac}(c). This is done by breaking the word-like patterns in $\theta^x_{i-1}$ into subword-like patterns, and then reconstructing new word-like patterns based on $W_i$. Those segments of several consecutive subword-like patterns appearing frequent enough and with high enough right and left context variation are taken as word-like patterns. This can be achieved by constructing an efficient data structure called PAT-Tree using the labels $W_i$\cite{pat}. In this way, the lexicon $\theta_i^x$ can be updated significantly in each iteration. This updated lexicon $\theta^x_i$ is then used in free-word decoding to produce the labels $W_{i+1}$. The whole process is completed when there is no significant difference between $W_i$ and $W_{i+1}$. This gives the automatically discovered linguistic structure $\theta=\{\theta^a,\theta^x,\theta^l\}$, where $\theta^l$ is trained from the final version of $W_{i+1}$.

\section{EXPERIMENTS}
\label{sec:typestyle}

\subsection{Experimental Setup}
The proposed approach was tested in the preliminary experiments performed on a corpus of Mandarin broadcast news collected in Taiwan in 2001 with length of 4 hours including 5034 utterances. The HMMs used for each sub-word like pattern had 13 states, each with only 1 Gaussian component. This configuration was selected due to the assumption that the subword-like patterns of interest should describe more signal trajectory variation and less acoustic variation. Signal segments with larger acoustic variation should be classified as different patterns.  The final linguistic structure including all patterns, models and parameters was obtained by performing 30 iterations in each stage (I)(II)(III)
in Fig. \ref{fig:FP} on the entire corpus.

\subsection{Initial Observations and Analysis}

\begin{figure}[tbh]
    \centering
    \includegraphics[width=0.45\textwidth]{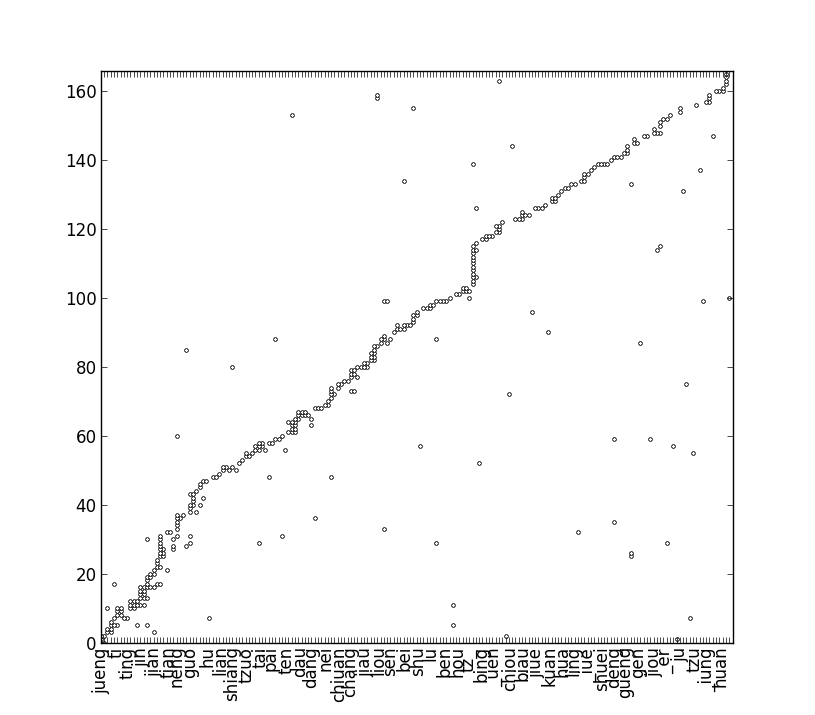}
    \caption{Mapping relation between the discovered subword-like patterns and Mandarin syllables. Only pairs with 35 or more occurrence are shown, and the average co-occurrence mapping for all circles in the figure is 331.}
    \label{fig:BP}
\end{figure}

It is interesting that almost all the 208 subword-like patterns obtained here roughly correspond to Mandarin syllables (each Chinese character is pronounced as a Mandarin syllable). A global view of the exact mapping relation from the 208 subword-like patterns to the total of 399 Mandarin syllables manually labelled for the corpus is shown in Fig. \ref{fig:BP}. The Mandarin syllables on the horizontal scale of the figure have been sorted according to acoustic similarity (only a quarter of them are explicitly printed due to limited space). Every circle here represents 35 or more subword-like patterns on the vertical scale whose central feature frame belonged to the Mandarin syllable in the horizontal scale. This figure implied a very-close-to one-to-one mapping relation with some fuzziness around neighbouring syllables with similar acoustic behaviour. The 362 word-like pattern obtained corresponded to roughly 154 frequently occurring multi-syllable words and 208 monosyllables (or mono-subword-like patterns). Those words occurring not frequently enough couldn't be discovered and as a result were represented as one to several mono-subword-like patterns. 

\begin{figure}[tbh]
    \centering
    \includegraphics[width=0.45\textwidth]{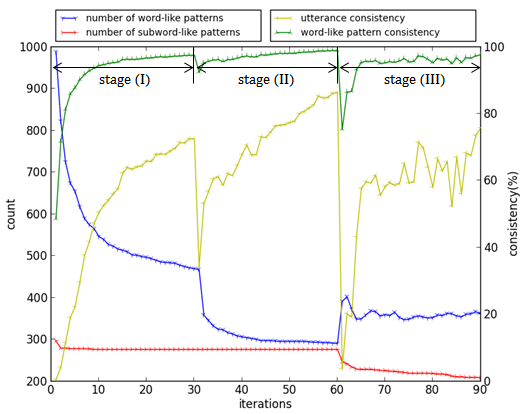}
    \caption{ Number of subword-like patterns, lexicon size for word-like patterns (left) and consistency between $W_i$ and $W_{i+1}$ in terms of word-like patterns and utterances (right) as functions of iterations. The transition from stage(I) to stage(II) and stage(II) to stage(III) happened at iteration 30 and 60 respectively.   
    }
    \label{fig:4line}
\end{figure}

Fig. \ref{fig:4line} further illustrates how the number of subword-like patterns, lexicon size of word-like patterns, the consistency between $W_{i-1}$ and $W_{i}$ at word-like pattern level and utterance level changed with respect to iterations. In a global perspective, lexicon size of word-like patterns dropped in the stages (I) and (II), and jumped and oscillated in stage (III). Although most word-like patterns in stage (I) did not survive by the end of stage (II), the main purpose of them was to provide some context guidance for the training of subword-like HMMs.

\subsection{Justification of the Initialization and Iterative Stages}
We performed further tests with configurations slightly different from the proposed approach on a subset of 942 utterances out of the 5034 in the tested corpus. We evaluated the syllable accuracy by mapping every discovered subword-like pattern to a corresponding Mandarin syllable (as was done in Fig. \ref{fig:BP}) for each configuration considered. In the first part, we initialized $W_0$ with 3 different methods and then applied 50 iterations of stage (I) only. The three methods are (1) the proposed two-level top-down labelling started with word-like segments, (2) subword initialization with only watershed transform, but without higher level word-like segments, (3) same as (2) but without k-means clustering, with same number of subword-like pattern IDs randomly assigned to each subword-like segment. The main difference between methods (1)(2) was the two-level pattern  structure. Method (1) brought us halfway through the proposed approach (initialization and stage (I)) producing two-level patterns, while method (2) was similar to the unsupervised initialization methods used previously with one-level patterns only \cite{jansen1}\cite{jansen2}. The results are in the left half of Table \ref{tab:ASR}. Although method (1) was only 1.03\% better than method (2), the patterns obtained with method (1) manual auditing tests suggest that the improvement is non-trivial. This verified the word-like pattern constraints were useful in the acoustic optimization process. The random ID assignments without clustering in method (3) also offered relatively high accuracy. This implied the acoustic optimization iterations in stage (I) was quite helpful. 

In the second part, we initialized $W_0$ with the two-layered method then applied 3 different iteration sequences:
%(a)30 iterations in stage (I) followed by 20 in stage (II), (b) 50 in stage(I), (3) 50 in stage (II).
%(a)$I_a=30$,$I_l=20$,$I_x=0$, (2)$I_a=50$,$I_l=0$,$I_x=0$, (3) $I_a=0$,$I_l=50$,$I_x=0$.
(1) $(I_a,I_l,$ $I_x)$ = $(30,20,0)$, (2) $(I_a,I_l,I_x)$ = $(50,0,0)$, (3) $(I_a,I_l,I_x)$ = $(0,50,0)$.
Method (1) brought us halfway through the proposed approach wile method (3) was actually the intuitive joint optimization of both acoustic and linguistic parameters similar to previously proposed approaches \cite{gish1}\cite{gish2}. The results are in the right half of Table \ref{tab:ASR}. The proposed method (1) was 2.37\% better than the joint optimization method (3). The proposed method (1) was also better than the applying method (2) alone, which implies that the transition was the source of improvement. This verified that gradually learning later after layer yielded more reliable results. The benefits of the lexical optimization in stage (III), on the other hand, are better observed in a companion paper on semantic retrieval of spoken content also submitted to ICASSP 2013\cite{lee}, since the word-like patterns carried semantics. 

\begin{table}[t]
\centering
\begin{tabular}{|p{1.8cm}|p{1cm}|p{2.8cm}|p{1cm}|} 
\hline
\multicolumn{2}{|c|}{(A)Initialization methods}&\multicolumn{2}{|c|}{(B)Iteration methods} \\
%a&a&g&a\\
\hline
%word-like unit&38.96 & transition &39.45  \\ 
(1)Two-level &38.96\%& (1)($I_a$,$I_l$,$I_x$)=(30,20,0) &39.45\% \\ 
\hline
%subword unit       &37.93 & acoustic   &38.96\\ 
(2)One-level       &37.93\%&  (2)($I_a$,$I_l$,$I_x$)=(50,0,0)   &38.96\%\\ 
\hline
%random      &35.76 & linguistic &37.08\\
(3)Random      &35.76\%&  (3)($I_a$,$I_l$,$I_x$)=(0,50,0) &37.08\%\\
\hline
\end{tabular}
\caption{ASR accuracy of unsupervised transcription translated by string replacement with most probable assignment}
\label{tab:ASR}
\end{table}

\subsection{Spoken Term Detection}
We also applied the discovered patterns on a task of spoken term detection \cite{std2}-\cite{std7} and compared to a set of Mandarin syllable models trained on a manually annotated corpus of 24.5 hours of Mandarin Broadcast News with a trigram for 72k vocabulary used in recognition.  The performance of the supervised HMMs serves as an upper bound for the performance of our unsupervised HMMs. We tested the performance of the supervised and unsupervised models under the same scenario. 
%The query set consisted of 52 text entities of countries, organizations and political leaders decoded from their corresponding utterances. 
The query set consisted of 52 name entities of countries, organizations and political leaders. For each query, we decoded their corresponding utterances in the corpus and selected the most frequent HMM sequence to represent each query (equivalent to query by one example of the best query utterance).
Syllable HMMs were used for the supervised case, and subword-like pattern HMMs were used for the unsupervised case. 
This query HMM sequence was then compared with the HMM sequences of all utterances in the corpus for evaluation of the relevance scores for retrieval.
%The result we get under this setting is actually the optimal retrieval performance we can get with one example query.
We first computed offline the distance between each pair of two HMMs. The distance between two HMMs was defined to be the DTW-distance between the two state sequences. One state in a HMM can be matched with several states in another HMM and vice versa. The distance metric used for DTW was the KL-divergence between the two Gaussian mixtures \cite{dist}. We then calculate the distance between the query HMM sequences and corpus HMM sequences online. The distance between two HMM sequences was defined to be the sum of distances for matched pairs of models for the two sequences. Since most computation was done offline, this method was as fast as text information retrieval.

\begin{figure}[tbh]
    \centering
    \includegraphics[width=0.5\textwidth]{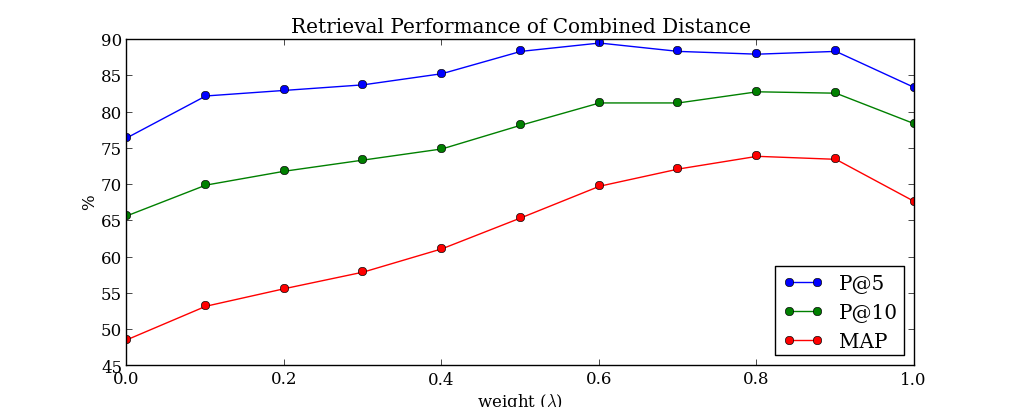}
    \caption{ The spoken term detection performance based on the weighted sum of unsupervised(left) and supervised(right) distance metrics.    }
    \label{fig:weight}
\end{figure}
We took the weighted sum of the supervised distance $d_s$ and unsupervised distance $d_u$, and performed spoken term detection based on the combined distance $d_\lambda = \lambda \times d_s + (1-\lambda) \times d_u$. The results in Fig. \ref{fig:weight} show that reasonable detection performance was achieved for the unsupervised model on its own ($\lambda = 0$). More importantly, the combined distance can yield better results in all the three measures than using only supervised or unsupervised distances. This implies that the proposed method has successfully harvested information directly from the data that was lost during recognition with the supervised models. In other words, the proposed method not only performs reasonably well on its own, but it is also complimentary to standard supervised ASR systems. 

%\begin{table}[t]
%\centering
%\begin{tabular}{|l|c|c|c|c|c|} 
%\hline
%Model(\%) & P@5 & P@10 & MAP \\
%\hline
%(A)Supervised Syllables &83.46 & 78.46& 67.87\\
%%(A)Supervised Syllables &82.3 & 79.2& 65.3\\
%%Chinese Monophone &83.46 & 78.46& 67.87\\
%\hline
%(B)Unsupervised Patterns & 76.54 & 65.77 & 48.66\\
%%(B)Unsupervised Patterns & 86.9 & 74.2 & 53.7\\
%%Proposed Model & 76.15 & 65.19 & 47.66\\
%\hline
%%(C)Combined (Arithmetic Mean) &88.46 & 78.27 & 65.48\\
%%\hline
%(C)Combined (Geometric Mean) &81.92  & 79.04 & 72.28\\
%\hline
%\end{tabular}
%\caption{Spoken Term Detection performance comparison}
%\label{tab:CHN}
%\end{table}

\section{CONCLUSION}
\label{sec:prior}

This work presents an approach for unsupervised discovery of linguistic structure including two-level acoustic patterns from a corpus. The main difference from similar approaches proposed earlier \cite{jansen1}\cite{glass}\cite{gish1}\cite{gish2}\cite{unsup2}\cite{unsup3}\cite{chan1} lies in the two-level acoustic patterns and the layer-after-layer gradual learning of the model parameters with cascaded stages of iterative optimization. Although some earlier approaches \cite{jansen1} also took hierarchical knowledge into consideration, our work used 13-state single Gaussian HMMs as compared to the conventional HMMs with smaller number of states and multi-Gaussian \cite{jansen1}\cite{glass}\cite{gish1}\cite{gish2} to model the trajectories of acoustic patterns with less acoustic variation. 
The preliminary experiment on spoken term detection on subword-like pattern sequences indicated that the proposed system is complimentary to existing ASR systems. A more complete experiment on spoken term detection in a companion paper submitted to ICASSP 2013 \cite{chan3} demonstrates how our model can outperform the segmental DTW approach.
%A more advanced version of the experiment comparing it to the baseline DTW could be found in a companion paper \cite{chan3}.
Also, the second level of word-like patterns are aimed to capture some semantic features in the acoustic signal, which can be verified in a companion paper on Semantic retrieval of spoken content also submitted to ICASSP 2013 \cite{lee}.

\bibliographystyle{IEEEbib}
\bibliography{myrefs}

\end{document}